# Evaluating Variable Length Markov Chain Models for Analysis of User Web Navigation Sessions

José Borges *

School of Engineering,

University of Porto

R. Dr. Roberto Frias, 4200 - Porto, Portugal

jlborges@fe.up.pt

Mark Levene

School of Computer Science and Information Systems,

Birkbeck, University of London

Malet Street, London WC1E 7HX, U.K.

mark@dcs.bbk.ac.uk

March 26, 2018

---

*Supported by the POSC/EIA/58367/2004/Site-o-Matic Project (Fundação Ciência e Tecnologia) co-financed by FEDER.

**Abstract**

Markov models have been widely used to represent and analyse user web navigation data. In previous work we have proposed a method to dynamically extend the order of a Markov chain model and a complimentary method for assessing the predictive power of such a variable length Markov chain. Herein, we review these two methods and propose a novel method for measuring the ability of a variable length Markov model to summarise user web navigation sessions up to a given length. While the summarisation ability of a model is important to enable the identification of user navigation patterns, the ability to make predictions is important in order to foresee the next link choice of a user after following a given trail so as, for example, to personalise a web site. We present an extensive experimental evaluation providing strong evidence that prediction accuracy increases linearly with summarisation ability.

# 1 Introduction

Web usage mining has been defined as the research field focused on developing techniques to model and study users' web navigation data [Mob04]. Data characterising user web navigation sessions can be collected from server log files or from a web browser plug-in aimed at recording the user navigation options. From the collected data it is possible to reconstruct users' web navigation sessions [SMBN03], where a session, also referred to as a *trail*, consists of a sequence of web pages viewed by a user within a given time window.

Several models have been proposed for modelling user web data. Schechter et al. [SKS98] utilised a tree-based data structure that represents the collection of paths inferred from the log data to predict the next page access. Dongshan and Junyi [DJ02] proposed a hybrid-order tree-like Markov model, which provides good scalability and high coverage of the state-space, also to predict the next page access. Chen and Zhang [CZ03] utilised a PPM (Prediction by Partial Match) forest that restricts the roots to popular nodes; assuming that most user sessions start with popular pages, they reduce the model complexity by eliminating branches having a non-popular page as their root. [DK04] proposed a technique that builds $k^{th}-order$ Markov models and combines them to include the highest order model covering each state; they also develop techniques to eliminate states and reduce the model complexity. In [EVK05] the authors propose a method that incorporates link analysis, such as the pagerank measure, into a Markov model in order to provide web path recommendations.

In [BL00] we have proposed a first-order Markov model for a collection of user navigation sessions, and more recently we have extended the method to represent higher-order conditional probabilities by making use of a cloning operation [BL05a, BL05b]. In addition, we have proposed a method to evaluate the *predictive power* of a model that takes into account a variable length history when estimating the probability of the next link choice of a user, given his/her navigation trail [BL06]. (We review these models in subsections 2.1 and 2.2.) Herein, we propose a new method to measure the *summarisation ability* of a model, by which we mean the ability of a variable length Markov model to summarise user web navigation sessions up to a given length. Briefly, from a collection of user web navigation trails a variable length Markov model, up to order $n$, is built, and an algorithm is devised to induce the user trails having a high probability of being traversed. The set of trails is then ranked, from the most to least probable, and the resulting ranking is compared with the ranking induced from input trails of length $n$, i.e. from the $n$-gram frequency count of the input trails. Thus, if a model accurately represents the collection of input navigation trails of length $n$, the two rankings should be identical. Note that while it is clear that a $n$-order Markov model accurately represents trails shorter or equal to $n$, it is important to understand how well such a model represents trails longer than $n$. To make the comparison between the rankings we use the Spearman footrule [FKS03] and percentage overlap metrics. We present the results of an extensive experimental evaluation conducted on three real world data sets, which provides strong evidence that the measure of predictive power increases linearly with the measure for summarisation ability.

Understanding the behaviour of web site visitors navigating through the site is an important step in the process of improving the quality of service of the site. In our opinion Markov models are well suited for modelling user web navigation data because they are compact, simple to understand and motivate, expressive and based on a well established theory. Commercial tools for log data analysis usually discard the information concerning the order in which page views occurred in a session and the same can be said about the techniques using association rules methods. On the other hand, variable length Markov chain models provide the probability of the next link chosen when viewing a web page, while taking into account the trail followed to reach that page.

Our measure of the summarisation ability of the model answers a question we have often been asked about the adequacy of Markov models in representing user web trails. The measure tells us how accurate is the Markov model representing the users' trails, and, thus, when having a precise Markov representation we are justified in using methods that extract the most probable trails from the model, see [BL04]. Such trail patterns can, for example, provide guidelines for improving the static structure of a web site. Moreover, Markov models can also be used to predict the user's next navigation step within the site, as is shown in Section 2.4. Such methods can be also used for web site personalisation by adapting the web page presentation to the profile of the current user. Thus, the summarisation ability of the model is important for enabling the identification of user navigation patterns, and the prediction ability is important for foreseeing the next link choice of a user after following a given trail.

We stress that there has been previous research on evaluating the ability of a Markov model to predict the next link choice of a user, [DK04, BL06], however, to the best of our knowledge, there is a lack of publications on evaluating the ability of a Markov model to represent user sessions up to a given history length. Moreover, as far as we know, the relationship between summarisation ability and prediction power has not been studied before in the context of web mining.

The rest of the paper is organised as follows. In Section 2 we introduce the variable length Markov chain methods we make use of. More specifically, in subsection 2.1 we introduce our method for building a first-order model from a collection of user sessions, in subsection 2.2 we introduce our extension of the method to higher-order conditional probabilities, in subsection 2.3 we present our new method for measuring the model's summarisation ability, and in subsection 2.4 we introduce our method for measuring the model's predictive power. In Section 3 we present an experimental evaluation of these methods, and, finally, in Section 4 we give our concluding remarks.

## 2  Variable Length Markov Chains Methods

A user navigation session within a web site can be represented by the sequence of pages requested by the user. First-order Markov models have been widely used to model a collection of user sessions. In such a model each web page in the site corresponds to a state in a first-order Markov model, and each pair of pages viewed in sequence corresponds to a state transition in the model. Each transition probability is estimated by the ratio of the number of times the transition was traversed and the number of times the first state in the pair was visited. Usually, an artificial state is appended to every navigation session to denote the start and finish of the session.

A first-order Markov model is a compact way of representing a collection of sessions but in most cases its accuracy is low [JPT03], which is why extensions to higher-order Markov models are necessary to improve the accuracy. In a (non-variable) higher-order Markov model a state corresponds to a fixed sequence of pages, and transition between two states represents a higher-order conditional probability [BL00]. For example, in a second-order model each state corresponds to a sequence of two page views. The serious drawback of fixed higher-order Markov models is their exponentially large state space compared to lower-order models.

A *variable length Markov chain* (VLMC) is a model extension that allows variable length history to be captured [Bej04]. In [BL05b] we proposed a method that transforms a first-order model into a VLMC so that each transition probability between two states takes into account the path a user followed to reach the first state prior to choosing the outlink corresponding to the transition to the second state. The method makes use of state cloning (where states are duplicated to distinguish between different paths leading to the same state) together with a clustering technique that separates paths revealing differences in their conditional probabilities. We will now introduce our method by means of an illustrative example; we refer readers looking for a formal description of the method to [BL05b].

## 2.1 First-Order Model Construction

Figure 1 (a) shows an example of a collection of user navigation sessions. We let a session start and finish at an artificial state and, for each session, we give the number of times the corresponding sequence of pages was traversed by a user (freq.). Figure 1 (b) presents the first-order model for these sessions. There is a state corresponding to each web page and a link connecting every two pages viewed in sequence. For each state that corresponds to a web page we give the page identifier and the number of times the page was viewed divided by the total number of page views. This ratio is an estimate of the probability of a user choosing the corresponding page from the set of all pages in the web site. For example, page 4 has 22 page views from a total of 101 page views. For each link we indicate the proportion of times it was followed after viewing the anchor page of the transition. For example, page 5 was viewed 14 times, 5 of which were at the beginning of a navigation session (the weight of the link from the artificial state, S, indicates the number of sessions that started in that page). After viewing page 5 the user moved to page 6 in 7 of the 14 times and terminated the session 7 times. According to the model, the probability estimate of a trail is defined as the product of the probability of the first state in the trail (i.e. the initial probability) and the probabilities of the traversed links (i.e. the transition probabilities). For example, the probability estimate for trail (3,4) is $21/101 \cdot 10/21 = 0.099$ and for trail (1,3,5) is $11/101 \cdot 9/12 \cdot 9/21 = 0.035$.

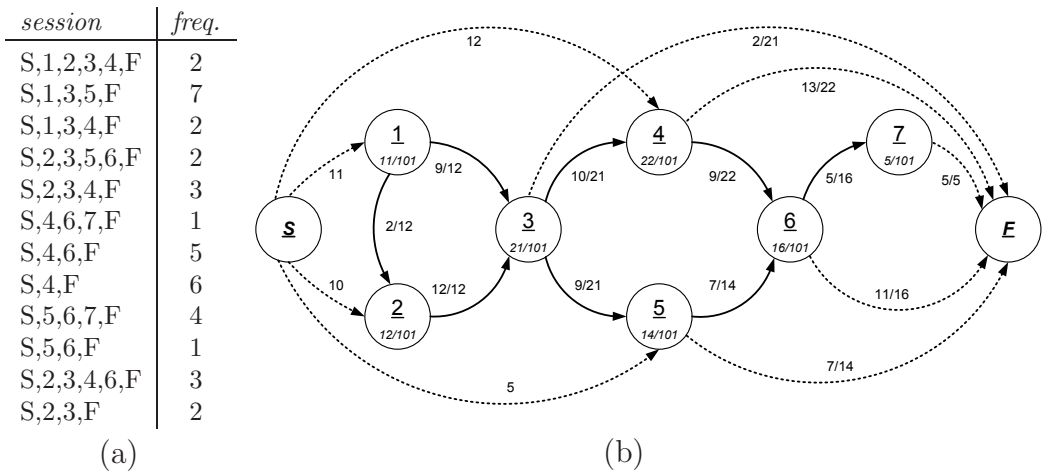

Figure 1: An example of a collection of sessions (a) and the corresponding first-order model (b)

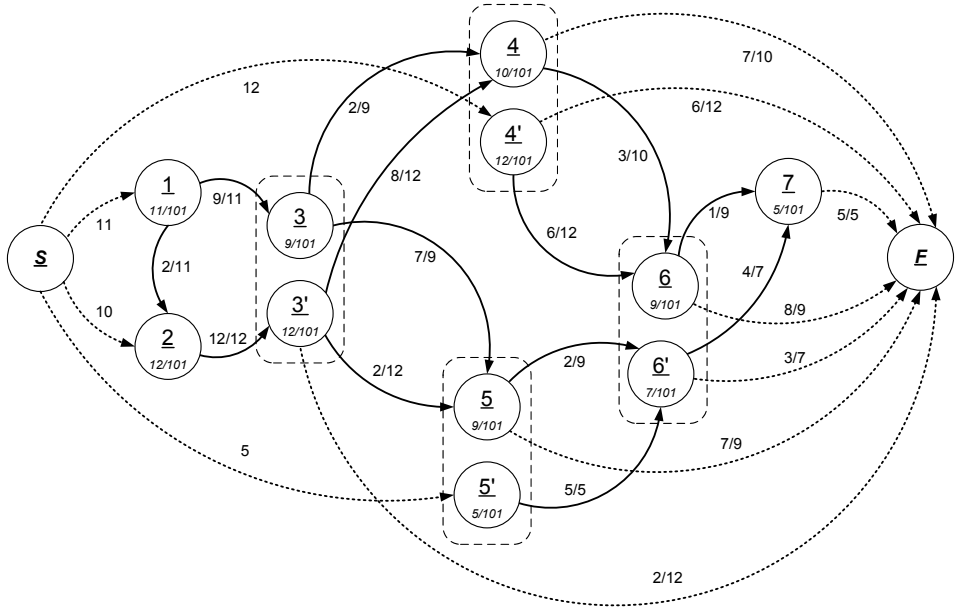

Figure 2: The second-order model for the example given in Figure 1

## 2.2 Higher-Order Model Construction

The first-order model does not accurately represent all second-order conditional probabilities. For example, according to the input data the sequence $(1, 3)$ was followed 9 times, i.e. $\#(1, 3) = 9$, and sequence $(1, 3, 4)$ was followed twice, i.e. $\#(1, 3, 4) = 2$. Therefore, the second-order conditional probability estimate for viewing page 4 after viewing 1 and 3 in sequence is $p(4|1, 3) = \#(1, 3, 4)/\#(1, 3) = 2/9$. The error of a first-order model in representing second-order probabilities can be measured by the absolute difference between a first-order probability and the corresponding second-order probability. For example, when assessing the accuracy of state 3 we have that $|p(4|1, 3) - p(4|3)| = |2/9 - 10/21| = 0.254$ and $|p(4|2, 3) - p(4|3)| = |8/12 - 10/21| = 0.190$ meaning that state 3 is not ac-curately representing second-order conditional probabilities. To address this problem the accuracy of the transition probabilities can be increased by separating the in-paths to the state corresponding to the different conditional probabilities. In this example increased accuracy can be achieved by cloning state 3 (i.e. creating a duplicate state 3') and redirecting the link (2,3) to state 3'. The weights of the outlinks from states 3 and 3' are updated according to the number of times the sequence of three states was followed. For example, since $\#(1,3,4) = 2$ and $\#(1,3,5) = 7$ in the second-order model, the weight of the link (3,4) is 2 and the weight of (3,5) is 7. (Note that, according to the input data, no session terminates at page 3 when the user has navigated to it from page 1.) The same method is applied to update the outlinks from the clone state 3'. Figure 2 shows the second-order model corresponding to the collection of sessions given in our example after four states were cloned in order to accurately represent all second-order conditional probabilities.

In the extended model given in Figure 2 all the outlinks represent the correct second-order probabilities estimates. The probability estimate of the trail (1,3,5) is now $11/101 \cdot 9/11 \cdot 7/9 = 0.069$. The probability estimate for trail (3,4) is computed as $(9/101 \cdot 2/9) + (12/101 \cdot 8/12) = 0.099$ which is equal to the probability estimate given by the first-order model, noting that state 3' is a clone of 3. Therefore, the second-order model accurately models the conditional second-order probabilities estimates while keeping the correct first-order probability estimates.

In order to provide control over the number of additional states created by the method we make use of a parameter, $\gamma$, that sets the highest admissible difference between a first-order and the corresponding second-order probability estimate. In a first-order model, when assessing the accuracy of a state the state is cloned if there is a second-order probability whose difference to the corresponding first-order probability is greater than $\gamma$. Alternatively, we interpret $\gamma$ as a threshold for the average difference between the first-order and the corresponding second-order probabilities for a given state. In the latter, the state is cloned if the average difference between first and second-order conditional probabilities surpasses $\gamma$. Moreover, if we set $\gamma > 0$ and the state has three or more inlinks we make use of a clustering algorithm to identify inlinks inducing identical conditional probabilities. When $\gamma$ is measuring the maximum probability of divergence, we will denote it by $\gamma_m$ and when it is measuring the average probability divergence, we will denote it by $\gamma_a$.

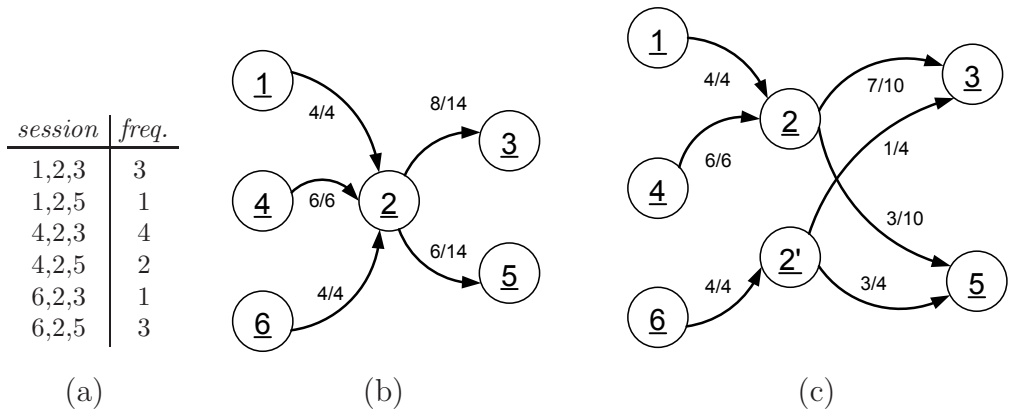

Figure 3: An example of a collection of sessions (a) and the corresponding first (b) and second-order models (c)

Figure 3 presents an example illustrating the role of the threshold accuracy parameter. In (a) a collection of sessions is given, in (b) the corre-sponding first-order model is shown, and in (c) the second-order model that results from creating one additional clone (states S and F are omitted for the sake of simplicity) is shown. In Table 1 the accuracy of both models is measured. The conditional probabilities estimated from the input data are given together with the maximum and average error attained by the first and second-order models when representing the conditional probabilities.

| probability | input data | 1st order | | 2nd order | |
|---|---|---|---|---|---|
| | prob. | prob. | diff. | prob | diff |
| $p(3\|1,2)$ | 3/4 | 8/14 | 0.18 | 7/10 | 0.05 |
| $p(5\|1,2)$ | 1/4 | 6/14 | 0.18 | 3/10 | 0.05 |
| $p(3\|4,2)$ | 4/6 | 8/14 | 0.10 | 7/10 | 0.03 |
| $p(5\|4,2)$ | 2/6 | 6/14 | 0.10 | 3/10 | 0.20 |
| $p(3\|6,2)$ | 1/4 | 8/14 | 0.32 | 1/4 | 0.00 |
| $p(5\|6,2)$ | 3/4 | 6/14 | 0.32 | 3/4 | 0.00 |
| | max | | 0.32 | | 0.20 |
| | avg | | 0.20 | | 0.06 |

Table 1: The conditional probability accuracy assessment for the models given in Figure 3

To create the second-order model, the method uses a clustering algorithm to identify which inlinks induce similar conditional probabilities, in this case links (1,2) and (4,2). Such links are assigned to the same clone. Depending on the value of the accuracy threshold, $\gamma$, the method decides if it is necessary to create an additional clone in order to increase the model's accuracy. As referred to above $\gamma$ can be set to regulate the maximum error ($\gamma_m$) or, alternatively, the average error ($\gamma_a$). In the example, if $\gamma$ was set to regulate the average error, then for $\gamma_a = 0.05$ the method would force the creation of one more clone and for $\gamma_a = 0.07$ the state is considered accurate.

The method to extend a model to third and higher orders is identical. $N$-order conditional probability estimates are compared to the corresponding lower order estimates, and the cloning method is applied to states that do not accurately represent the $n$-order estimates in order to separate the $n$-state length in-paths to the state being cloned. A formal description of the method is given in [BL05b].

## 2.3 Evaluation of the Summarisation Ability of a Model

From the definition of a model it follows that a first-order model accurately represents the probability estimates of all trails of length 2, that is, trails that visit 2 pages. Similarly, a second-order model accurately represents the probability estimates for all trails of length up to 3, and so on for higher-order models. In practice we will only build models up to a fixed order, say $n$, and so we would like to measure the ability of this $n$-order model to provide probability estimates for trails that are longer than $n$; we call the output of such a measure the *summarisation ability* of a model.

We make use of two metrics to measure the summarisability of a model: (i) the Spearman footrule with a location parameter [FKS03], which measures the proximity between two $top\_m$ lists, and (ii) the overlap between two $top\_m$ lists. In the context of this paper each element of a list is a trail of a given length and the trails are order by probability.

The metrics are defined as follows. Given two $top\_m$ lists $L_1$ and $L_2$ each with $m$ elements, we let $L$ be the union of the elements in the two lists and the location parameter be $m+1$. In addition, we let $f(i)$ be a function that returns the ranking of any element $i \in L$ in $L_1$, and when $i \notin L_i$ we let $f(i) = m + 1$; $g(i)$ is the equivalent function for $L_2$.

The footrule metric is now defined as:

$$F(L_1, L_2) = 1 - \frac{\sum_{i \in L} |f(i) - g(i)|}{MAX},$$

where $MAX$ is a normalisation constant, which for a *top_m* list is $m \cdot (m+1)$ corresponding to the case when there is no overlap between the two lists, and we subtract 1 from the fraction since we are interested in proximity rather than distance. In addition, we use list overlap to measure the number of elements we were able to identify relative to a reference set, which is the ranking that is considered to be correct. Given, the two *top_m* lists $L_1$ and $L_2$ and assuming that $L_1$ is the reference set, the overlap is defined as the percentage of elements in the list, $L_2$, being assessed, which occur in the reference list, $L_1$. We note that while the overlap provides a simple measure of the $L_2$ quality, the footrule metric has the advantage of taking into account the relative ranking of the elements occurring in both lists.

We will now illustrate the method using an example. We will first measure the accuracy with which the first-order model, shown in Figure 1, represents the trails having length 3 that are given in the input data. First, from the collection of sessions we induce all sequences of three pages, the 3-grams, and rank them by frequency count. The *top_m* 3-grams, with $m = 5$, are presented in the first list, $L_1$, and will constitute the reference set. Second, we use a Breadth-First-Search (BFS) algorithm to infer the set of trails induced by the first-order models and each trail probability estimate; see [BL04] for details on the algorithm and its average linear time complexity. The *top_m* trails induced by the first-order model are presented in the second list, $L_2$. In order to ensure consistency between the rankings we lexicographically sort the 3-grams having the same frequency count and the trails having the same probability. Similarly, $L_3$ and $L_4$ are the lists of the *top_m* trails induced by the second and third-order models.

Table 2 presents the results for both the first and second-order models. For the first-order model, 3 of the *top_5* trails are in the *top_5* 3-grams, therefore, the overlap between $L_2$ and the reference set $L_1$ is $3/5 = 0.60$. The union of the two lists has 7 elements and, for each of its elements, we compute the rank absolute difference as given by the two lists. For example, $|f(2,3,4) - g(2,3,4)| = |1 - 3| = 2$ and $|f(1,3,5) - g(1,3,5)| = |3 - 6| = 3$. The footrule metric has the value $F(L_1, L_2) = 1 - 12/30$, which in this case is 0.60. As expected, the second-order model provides a trail ranking which is exactly the same as the one given by the 3-gram frequency counts.

| rank | $L_1$ 3-gram | freq | $L_2$ trail | prob | $L_3$ trail | prob |
|---|---|---|---|---|---|---|
| 1 | 2,3,4 | 8 | 4,6,F | 0.061 | 2,3,4 | 0.0792 |
| 2 | 4,6,F | 8 | 3,4,F | 0.059 | 4,6,F | 0.0792 |
| 3 | 1,3,5 | 7 | 2,3,4 | 0.057 | 1,3,5 | 0.0693 |
| 4 | 3,4,F | 7 | 2,3,5 | 0.051 | 3,4,F | 0.0693 |
| 5 | 3,5,F | 7 | 6,7,F | 0.050 | 3,5,F | 0.0693 |
| | | footrule | 0.60 | | 1.00 | |
| | | overlap | 0.60 | | 1.00 | |

Table 2: The ranking of 3-grams as given by the input data ($L_1$), the first-order model ($L_2$) and the second-order model ($L_3$)

Table 3 presents the results of the analysis for trails having length 4. As expected, the first-order model is less accurate for trails of length 4, and as result we obtain a lower value for both the footrule metric and the overlap. The first-order model is able to identify only 2 of the *top_5* trails and is not able to identify the most frequently traversed 4-gram. The second-order

|      | $L_1$   |      | $L_2$   |        | $L_3$   |        |
|------|---------|------|---------|--------|---------|--------|
| rank | 4-gram  | freq | trail   | prob   | trail   | prob   |
| 1    | 1,3,5,F | 7    | 2,3,4,F | 0.0334 | 2,3,4,F | 0.0554 |
| 2    | 2,3,4,F | 5    | 3,5,6,F | 0.0306 | 1,3,5,F | 0.0539 |
| 3    | 5,6,7,F | 4    | 3,4,6,F | 0.0278 | 5,6,7,F | 0.0396 |
| 4    | 2,3,4,6 | 3    | 4,6,7,F | 0.0278 | 3,4,6,F | 0.0264 |
| 5    | 3,4,6,F | 3    | 2,3,5,6 | 0.0255 | 2,3,4,6 | 0.0237 |
|      | footrule |     | 0.34    |        | 0.87    |        |
|      | overlap  |     | 0.40    |        | 1.00    |        |

Table 3: The ranking of 4-grams as given by the input data ($L_1$), the first-order model ($L_2$) and the second-order model ($L_3$)

trails fully overlaps with the 4-grams but only the trail (5,6,7,F) is attributed the correct rank.

In Table 4 we present the *top_3* results for models up to the third-order when analysing their ability to represent trails having length 5. (We consider only the *top_3*, since in the input data there are only 3 distinct 5-grams.) It is interesting to note that the first-order model identifies two of the top three trails, however, the missing trail is ranked in the eleventh place by the model (not shown in the table). As expected, the accuracy of the results improve as the order of the model increases.

|      | $L_1$     |      | $L_2$     |        | $L_3$     |        | $L_4$     |        |
|------|-----------|------|-----------|--------|-----------|--------|-----------|--------|
| rank | 5-gram    | freq | trail     | prob   | trail     | prob   | trail     | prob   |
| 1    | 2,3,4,6,F | 3    | 2,3,5,6,F | 0.0175 | 2,3,4,6,F | 0.0211 | 2,3,4,6,F | 0.0297 |
| 2    | 1,2,3,4,F | 2    | 2,3,4,6,F | 0.0159 | 3,5,6,7,F | 0.0113 | 2,3,5,6,F | 0.0198 |
| 3    | 2,3,5,6,F | 2    | 3,5,6,7,F | 0.0139 | 1,2,3,4,F | 0.0092 | 1,2,3,4,F | 0.0124 |
|      | footrule  |      | 0.50      |        | 0.67      |        | 0.83      |        |
|      | overlap   |      | 0.67      |        | 0.67      |        | 1.00      |        |

Table 4: The ranking of 5-grams as given by the input data and by models of up to third-order

## 2.4 Evaluation of the Predictive Power of a Model

In previous work we have presented a study focused on evaluating a model's ability to predict the last page of a session based on the preceding sequence of pages viewed, [BL06]. We will now introduce this method.

To assess a model's prediction ability we randomly split the set of input trails into a training set and a test set, and then induce the model from the training set. For each test trail of length $l$ we use its $l-1$ prefix to predict the last page in the trail, which we call the prediction target (tg). The model induced from the training set gives the set of reachable pages (rp) from the tip of the prefix, assuming that the user has followed the sequence of pages defined by the prefix. The reachable pages are ranked by probability, with the most probable page having $rank = 1$. In order to measure the prediction accuracy of the model, we let the *Absolute Error* (AE) measure the prediction strength by setting AE $= rank-1$, i.e. the closer AE is to zero the better the prediction is. The overall prediction accuracy metric is given by the *Mean Absolute Error* (MAE) which is defined as the sum of AE over all the test trails divided by the number of test trails.

We illustrate the method by assuming that the input trails given in Figure 1 (a) constitute the training set and that we have a test set composed of the following three trails $\{(1,3,5),(2,3,5,6),(1,3,5,6,7)\}$. According to the first-order model after following the prefix (1,3) there are three reachable pages (see Figure 1 (b)), the pages corresponding to states 4, 5 and F (the latter corresponds to terminating the navigation session). In this case, the prediction target (state 5) has rank 2 among the reachable pages resulting in an absolute error AE=2-1=1. According to the second-order model (see

Figure 2), after following (1,3) the most probable choice is the link to page 5. Therefore, for the first trail in the test set the second-order model provides a more accurate prediction than the first-order model. For the second test trail the opposite occurs, although the overall MAE metric for the example confirms that the second-order model provides better predictions; see Table 5 for the details.

| trail | tg | rp | prob. | rank | AE |
|---|---|---|---|---|---|
| 1,3,5 | 5 | 4 | 10/21 | 1 | |
| | | 5 | 9/21 | 2 | 1 |
| | | F | 2/21 | 3 | |
| 2,3,5,6 | 6 | 6 | 7/14 | 1 | 0 |
| | | F | 7/14 | 1 | |
| 1,3,5,6,7 | 7 | F | 11/16 | 1 | |
| | | 7 | 5/16 | 2 | 1 |
| | | | $MAE = 0.667$ | | |
| | | | (a) | | |

| trail | tg | rp | prob. | rank | AE |
|---|---|---|---|---|---|
| 1,3,5 | 5 | 5 | 7/9 | 1 | 0 |
| | | 4 | 2/9 | 2 | |
| 2,3,5,6 | 6 | F | 7/9 | 1 | |
| | | 6 | 2/9 | 2 | 1 |
| 1,3,5,6,7 | 7 | 7 | 4/7 | 1 | 0 |
| | | F | 3/7 | 2 | |
| | | | $MAE = 0.333$ | | |
| | | | (b) | | |

Table 5: The prediction results with the first-order (a) and second-order (b) models for a set of three test trails

## 3 Experimental Evaluation

We conducted experiments with three real data sets. The first data set (CS) was made available by the authors of [SMBN03]. It originates from the DePaul University CTI web site (`www.cs.depaul.edu`) and corresponds to two weeks of site usage during 2002; sessions were inferred using cookies. The second data set (MM) corresponds to two weeks of site usage from the Music Machines site (`machines.hyperreal.org`) during 1999 and was made available by the authors of [PE00]. The third data set (LTM) rep-resents a month of site usage from the London Transport Museum web site (www.ltmuseum.co.uk) during January 2003. In the CS data set the sessions were already identified, and for the other two data sets a session was defined as a sequence of requests from the same IP address with a time limit of 30 minutes between consecutive requests. Erroneous and image requests were eliminated from the data sets, although for the MM data set .jpg requests were left in, since in that specific site they correspond to page views. When preprocessing the data sets we set a session length limit of 15 requests, and, therefore, very long sessions were split into two or more shorter sessions.

Table 6 summarises the characteristics of the data sets. For each data set we indicate the number of pages occurring in the log file and the total number of requests recorded. We also give the total number of sessions derived from each data set and the number of sessions of length one ($l = 1$), two ($l = 2$) and three ($l = 3$); session length is measured by the number of requests a session is composed of.

| data set | pages | requests | sessions | $l=1$ | $l=2$ | $l=3$ |
|---|---|---|---|---|---|---|
| CS | 547 | 115448 | 24548 | 7148 | 3474 | 2202 |
| LTM | 1362 | 372434 | 47021 | 13489 | 3428 | 1893 |
| MM | 8237 | 303186 | 50192 | 12644 | 7891 | 4925 |

Table 6: Summary characteristics of the three real data sets used for the experimental evaluation

From a collection of sessions we infer the corresponding $n$-grams. An $n$-gram is defined as a sequence of $n$ consecutive requests. Figure 4 (a) shows the distribution of the $n$-gram frequency counts for the three data sets. For example, the frequency count of the 2-grams gives the number of distinct sequences of two pages that occur in the collection of sessions. Each user session was modified so that it starts and finishes at a fixed artificial page, resulting in sequences of at most seventeen requests as shown in the plot. As expected there is a higher variety of shorter sequences of pages implying that long sequences of page views are, generally, rare.

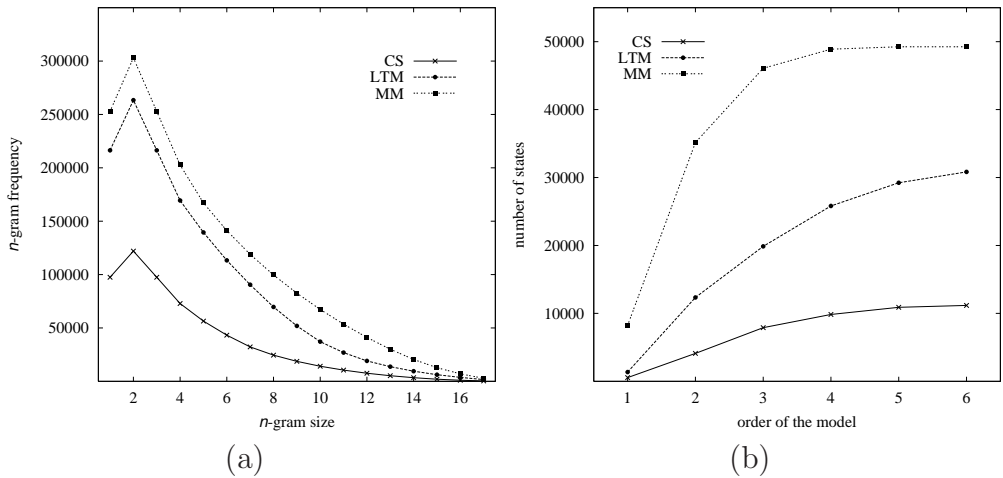

Figure 4: The distribution of the number of $n$-grams with each size (a) and the number of states with the model order (b) when $\gamma = 0$

From the collection of sessions a first-order model was inferred. This model was then evaluated for second and higher-order conditional probabilities and, if needed, a state was cloned to separate the in-paths due to differences in the conditional probabilities. As described above the $\gamma$ parameter sets the tolerance allowed on representing the conditional probabilities. In addition, there is a parameter that specifies the minimum number of times a page has to be requested in order to be considered for cloning, in these experiments we set $num\_visits \geq 30$. Figure 4 (b) shows the variation of the model number of states when the order of the model increases, while having the accuracy threshold set to $\gamma = 0$. The figure shows that there is a fast in-crease in the number of states for second and third order models. For higher orders the number of states increases at a slower rate, which is an indication that the gain in accuracy when using higher order models is smaller.

As discussed in Section 2.2, when $\gamma = 0$ a model of a given order accurately represents trails of length $\leq$ order $+$ 1 (the length of a trail is measure by the number of pages views). For example, a first-order model accurately represents the probability of a transition between any two states and a second-order model accurately represents the probability of trails composed of three consecutive states. In the following we make use of the method described in Section 2.3 to assess the ability of a model to represent long trails. Briefly, the method compares the ranking produced by the BFS algorithm that infers the trails' probability estimates with the ranking induced by the $n$-gram frequency count. When the model is accurate the two resulting rankings are identical.

For this purpose we consider three parameters: (i) the cut-point ($\lambda$), i.e. only trails having probability above $\lambda$ are considered, (ii) the maximum length of a trail induced by the algorithm ($mtl$), and (iii) the size of the ranked lists ($top\_m$). We consider two variations for the $mtl$ parameter: (i) a strict definition ($mtl_=$) that takes into account only trails with the specified length, and (ii) a non-strict definition ($mtl_{<=}$) that takes into account trails with length less than or equal to the specified limit. In (ii) we filter out subtrails, that is, trails that are a prefixes of longer trails whose probability is also above the cut-point since, by definition of trail probability, all trail prefixes have probability greater or equal to that of the full trail. As mentioned in Section 2.3, in order to compare the rankings we make use of two metrics:

(i) *overlap*, and (ii) Spearman's *footrule* metric. The overlap measures the percentage of trails that occur in both rankings, while the footrule takes into account not only the overlap but also the differences of rank ordering between the lists. We note that although the cut-point is essential to reduce the search space for the BFS algorithm, we will not emphasise it in our presentation, since the *mtl* is the parameter that directly influences the *top_m* ranking list. In fact, the cut-point was necessary for large values of *mtl* due to the very large number of trails to assess, however, its value did not have any impact on the resulting *top_m* ranking.

Figure 5 (a) shows, for the LTM data set, the footrule value for the *top_250* trails when using the strict trail length definition, $mtl_=$. The results show that a first-order model accurately ranks trails composed of one and two requests. For trails of length three the value of the footrule is 0.85 revealing a close to linear decrease for longer trails. We point out that the second-order model shows a footrule value of 0.92 for trails of length four. Figure 5 (b) shows the overlap measure whose results are very similar to the footrule metric. The overlap value shows a close to linear decrease as $mtl_=$ increases. It is interesting to note that a first-order model achieves an overlap of 0.50 for trails having length 5, which means that 50% of the *top_250* trails are identified in the list although probably not in the correct rank order. A second-order model achieves an overlap of 0.64 for trails of length 6. Similar results were obtained for the other two data sets.

Figure 6 shows, for the CS data set, the results when the non-strict definition of the trail length limit parameter ($mtl_{<=}$) is used. Again, the results given by the footrule and overlap metrics are very similar. In general, the non-strict definition gives better results for both the footrule and the overlap metrics, especially for longer trails. For example, for the CS data set and trails having length 5 the second-order model achieves footrule= 0.9 for the non-strict definition (as seen in Figure 6 (a)) and footrule= 0.78 for the strict definition. While the former is evaluating the ability to rank trails having length equal to or shorter than 5, the latter is assessing the ability of ranking trails of length 5 only. We note that, shorter trails are, in general, more probable and, when using the non-strict definition, if $mtl_{<=}$ increases from 4 to 5 very few additional trails with length 5 are included in the updated $top\_250$. Therefore, the strict definition for the $mtl$ parameter is more adequate for evaluating a model's ability to represent long trails.

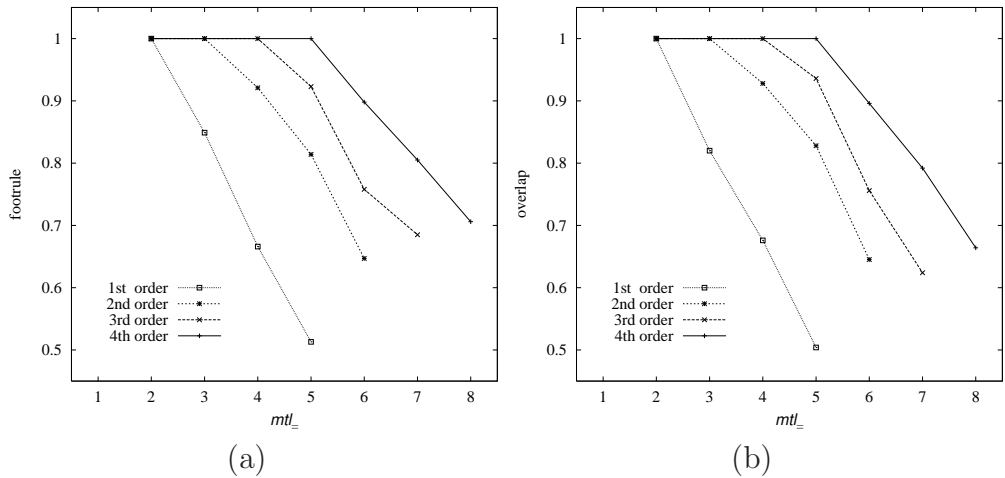

Figure 5: The footrule (a) and overlap (b) measures for the $top\_250$ trails on the LTM data set when using the maximum trail length strict definition ($mtl_=$)

Figure 7 (a) shows, for a third-order model of the CS data set, the variation of the footrule measure for different sizes of $n$ for the $top\_m$ ranking.

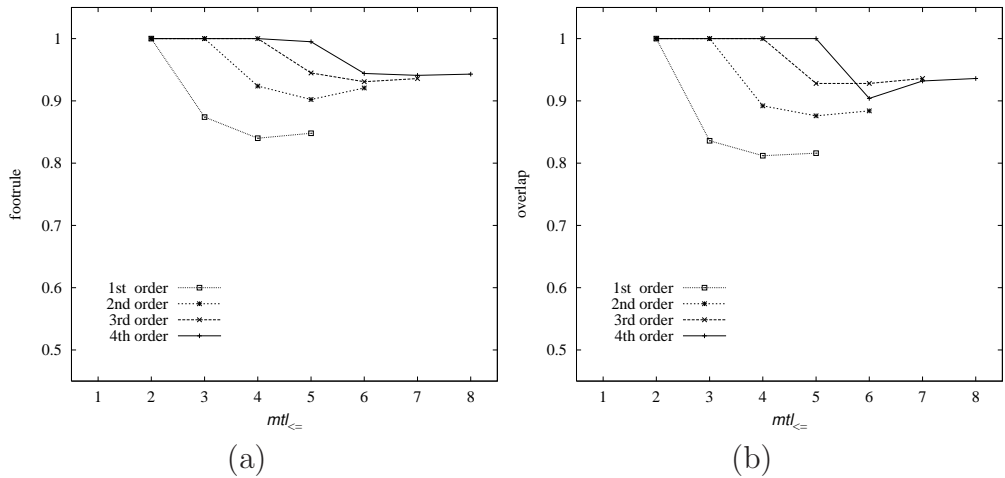

Figure 6: The footrule (a) and overlap (b) measures for the $top\_250$ trails on the CS data set when using the maximum trail non-strict definition ($mtl_{<=}$)

The footrule is smaller for larger $n$ but the differences are not very significant for trails of length up to 6. Similar results were obtained for the MM data set as is shown in Figure 7 (b). Therefore, from now on we restrict our analysis to the $top\_250$ rankings.

Figure 8 shows the variation of the number of states of the model with several values of the $\gamma$ parameter. For the results in (a) the $\gamma$ parameter measures the maximum divergence between a conditional probability and the corresponding lower order probability, in this case the parameter is denoted by $\gamma_m$. For the results in (b) $\gamma$ measures the average difference between the in-paths conditional probabilities and the corresponding link probabilities. In this case we will denote the parameter by $\gamma_a$. The results suggest that $\gamma_a$ is better since it provides more control over the resulting number of states in the model.

Figure 9 (a) shows the variation of the footrule measure with $\gamma_a$ for a third-order model on the CS data set taking the $top\_250$ rankings. It is interesting to note that the plotted lines are close to parallel, which indicates that the decrease in the footrule value for an increment of $mtl_=$ is close to being independent of the value of $\gamma_a$. Also, when $mtl_=$ is set to 5 and $\gamma_a$ increases from 0 to 0.1 the footrule decreases from 0.91 to 0.65, that is, it decreases by 29%. However, the number of states decrease from 7897 to 1011, which corresponds to a decrease of about 87%. Figure 9 (b) shows identical results for the LTM data set, but indicating the number of states obtained for each given $\gamma_a$ instead of the actual $\gamma_a$ value. Note that this presentation is essentially equivalent, since the $\gamma_a$ values have a direct effect on the number of states.

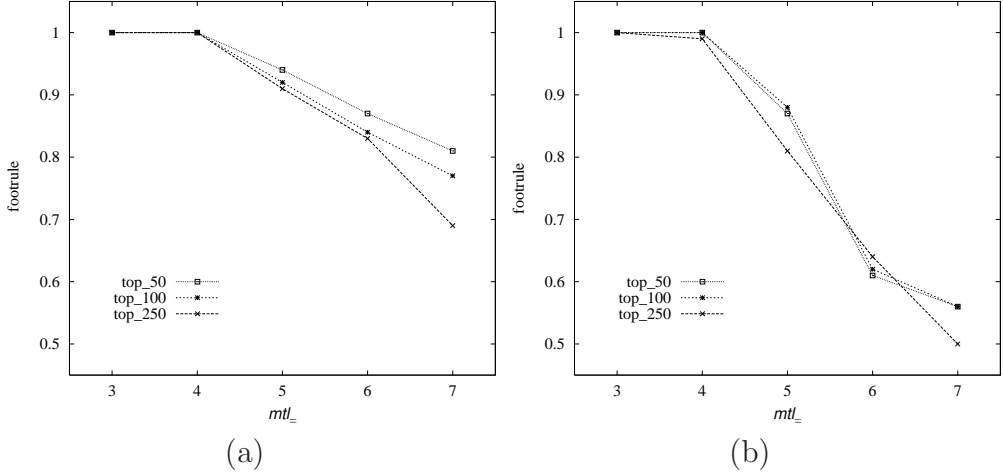

Figure 7: Analysis of the footrule measure sensitivity to the size of the $top\_m$ rankings for the CS data set (a) and the MM data set (b)

In the following we have decided to use the CS data set for the analysis of the predictive power of a model and its relationship to summarisation ability, since its sessions are more reliable than those inferred from the other data

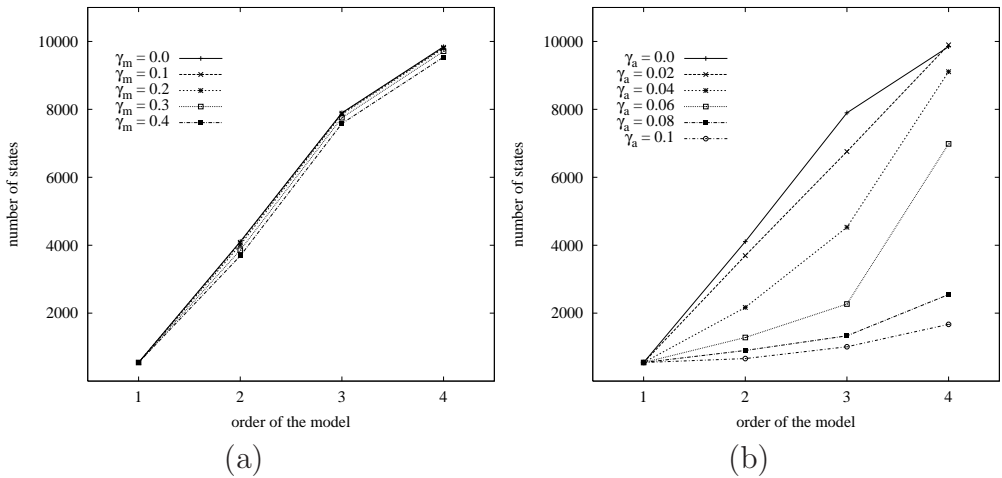

Figure 8: The variation of the number of states with the two versions of the accuracy threshold parameter, $\gamma_m$ (a) and $\gamma_a$ (b)

sets due to the use of cookies to identify the user.

We use cross validation in order to assess the predictive power of a model, i.e. we split the data set into $k$ folds and use $k-1$ folds as the training set and the remaining fold as the test set. In this experiments we will split the collection of trails in such a way that all trails in the $i$th subset temporally precede the trails in the $(i+1)$th subset. For $k = i$ we infer the model from all the trail subsets from 1 to $i$ and measure the prediction accuracy by using the $(i+1)$th subset as our test set. By splitting the data in this way we maintain the temporal order among the partitions. We make use of the MAE metric to measure the model's prediction accuracy. In addition, we define the complement of the MAE (MAE_c), which results from computing the rank of the prediction target relative to the last of the set of reachable pages. Since the sum of MAE and MAE_c is constant for a given $k$, we define the normalised measure, $st\_MAE = MAE/(MAE + MAE\_c)$. The normalised measure is interesting, since it takes into account the difficulty associated with the prediction, where a prediction is harder if it has more possible outcomes.

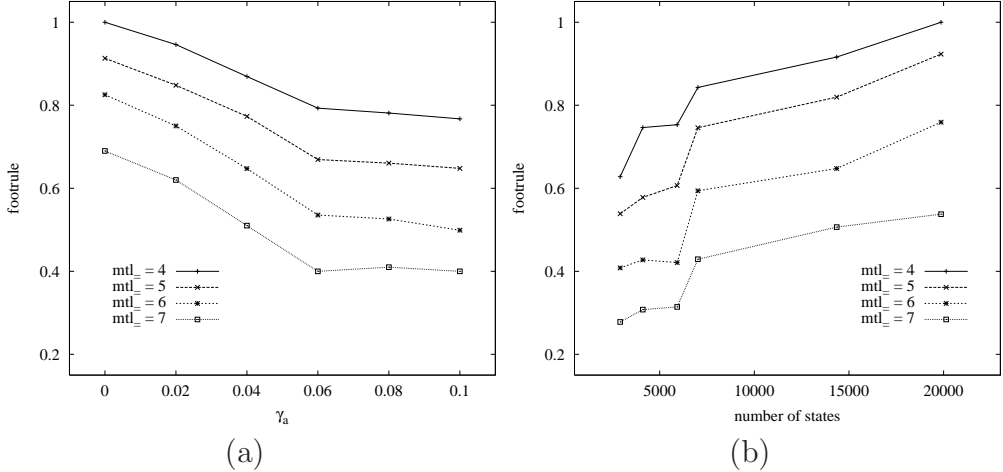

Figure 9: The variation of the footrule measure with $\gamma_a$ for third-order models on the CS (a) and LTM (b) data sets

|  | **k=2** | | | **k=3** | | | **k=4** | | |
|---|---|---|---|---|---|---|---|---|---|
| order | states | MAE | st_MAE | states | MAE | st_MAE | states | MAE | st_MAE |
| 1st | 411 | 4.415 | 0.099 | 464 | 5.227 | 0.105 | 511 | 4.944 | 0.083 |
| 2nd | 2315 | 3.298 | 0.074 | 3052 | 3.710 | 0.075 | 3631 | 3.576 | 0.060 |
| 3rd | 3727 | 3.087 | 0.069 | 5320 | 3.392 | 0.068 | 6661 | 3.290 | 0.055 |
| 4th | 4439 | 3.030 | 0.068 | 6472 | 3.296 | 0.066 | 8262 | 3.217 | 0.054 |

Table 7: The MAE and st_MAE for the temporal based 5-fold cross validation on the CS data set when $\gamma_a = 0.0$

Table 7 presents the results for models created with $\gamma_a = 0$. For all configurations of $k$ the prediction accuracy improves with the order of the model, but the gain when moving to the fourth-order model is relatively small. It is interesting to note that for first-order models both the MAE and st_MAE metrics increase from $k = 2$ to $k = 3$, but they decrease when moving to $k = 4$. We remind the reader that the higher $k$ is the larger the data set used to construct the model is, resulting in models with a larger number of states (as can be verified in Table 7) on which predictions are more difficult. Another aspect that should be taken into account is that when $k$ increases the sessions in the training set cover a larger period of time. For example, for $k = 4$ we are training the model with data from four periods of time and testing with the last period. In cases where the user's behaviour changes over time a model trained with data covering a larger period may have problems in generalising for future user behaviour.

| states | MAE | st_MAE |
|---|---|---|
| 512 | 4.787 | 0.081 |
| 4000 | 3.471 | 0.059 |
| 9533 | 3.090 | 0.052 |
| 13173 | 2.961 | 0.050 |

Table 8: The MAE and st_MAE on the CS data set for the non-temporal based 5-fold cross validation when $\gamma_a = 0.0$

In Table 8 present the results obtained with the standard 5-fold cross validation; we note that in [BL06] we used the standard cross validation rather than the temporal-based version used herein. When comparing the results we must take into account the fact that each value presented corresponds to the average result for the 5 splits. Moreover, the number of states is larger than that obtained from the temporal-based split, since we used $num\_visits \geq 15$, which leads to more states being cloned. However, overall, the results are very similar to those obtained with the temporal-based cross validation.

Table 9 gives the results obtained when $\gamma_a = 0.08$. As expected, when $\gamma_a$ increases the accuracy of the conditional probabilities decreases but we obtain models having a significantly smaller number of states. For a gain in the size of the model there is a cost in the loss of prediction accuracy. For example, for $k = 4$ folds a second-order model built with $\gamma_a = 0$ has 3631 states and when $\gamma_a = 0.08$ it has 829 states, that is, only 22.8% of the states are utilised in the latter case. However, the normalised error measure increases from 0.06 to 0.081, which corresponds to an increase of 31%.

|       | k=2 | | | k=3 | | | k=4 | | |
|-------|--------|-------|--------|--------|-------|--------|--------|-------|--------|
| order | states | MAE   | st_MAE | states | MAE   | st_MAE | states | MAE   | st_MAE |
| 1st   | 411    | 4.415 | 0.099  | 464    | 5.227 | 0.105  | 511    | 4.944 | 0.083  |
| 2nd   | 725    | 4.352 | 0.097  | 796    | 5.164 | 0.104  | 829    | 4.877 | 0.081  |
| 3rd   | 1152   | 4.173 | 0.093  | 1301   | 4.724 | 0.095  | 1226   | 4.579 | 0.076  |
| 4th   | 1823   | 3.808 | 0.085  | 2518   | 4.270 | 0.086  | 2356   | 4.181 | 0.070  |

Table 9: The MAE and st_MAE for the 5-fold cross validation on the CS data set when $\gamma_a = 0.08$

We now analyse the relationship between the footrule measure and the st_MAE measure. While the footrule measures the model's ability to summarise the information in a collection of trails, the st_MAE metric measures the model's ability to predict the next page view of a, possibly unseen, trail. To highlight the difference, summarisation represents the knowledge present in a collection of trails and prediction attempts to generalise this knowledge so that the outcome of future events can be predicted. In this set of experiments we set $mtl_=$ to 4 and the list sizes to the $top\_250$ trails.

Figure 10 (a) presents the results for a second-order model, with $\gamma_a$ set to the values in $\{0.00, 0.02, 0.04, 0.06, 0.08, 0.10\}$. Each plotted line corresponds to a cross validation partition and each point corresponds to a $\gamma_a$ value. When $\gamma_a$ increases the value of the footrule decreases and the value of st_MAE increases. When adjusting a regression line for $k = 2$, i.e for the 2-fold partition, we obtain 0.979 for the coefficient of determination, meaning that 98% of the footrule variation is explained by the variation of st_MAE. The other partitions give even higher values for the coefficient of determination.

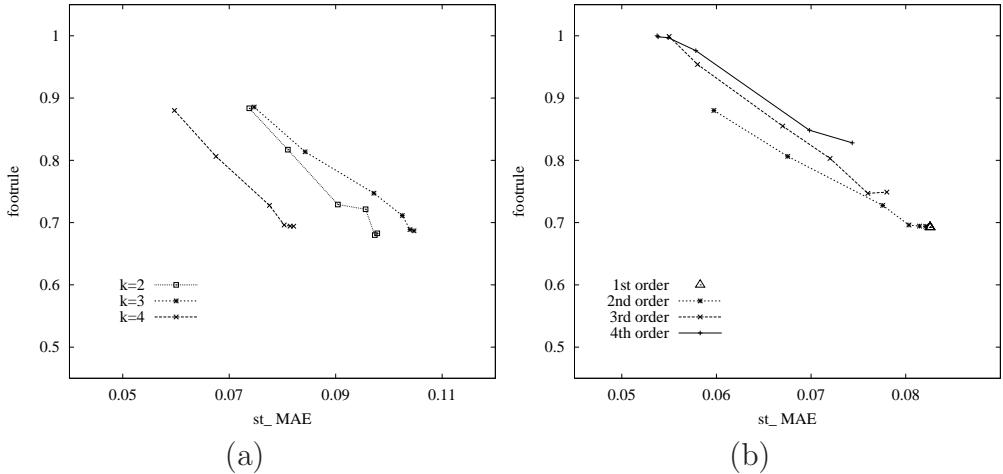

Figure 10: The relationship between the footrule measure and the st_MAE measure for second-order models while varying both $\gamma_a$ and the cross validation partition (a) and for the $k = 4$ cross validation partition while varying both $\gamma_a$ and the model order (b) on the CS data sets

Figure 10 (b) presents the results for $k = 4$, i.e. for the 4-fold partition, and for $mtl_=$ set to 4, while varying both the model order and the value of $\gamma_a$. The first-order model corresponds to a single point, since the $\gamma_a$ value has no effect in this case; moreover, it corresponds to the worst performance with respect to both measures. The regression line adjusted to the fourth-order results plot give 0.989 for the coefficient of determination (which is the lowest among the three plotted lines). Therefore, the results suggest that prediction accuracy improves linearly with summarisation ability.

Finally, we analyse the $top\_10$ trails extracted from the MM data set with models of different orders. While Table 10 presents the URLs occurring in the trails and provide an ID for each URL, Table 11 shows the 10 top ranked trails having length four that are inferred from first, second and third-order models, with $\gamma_a = 0.0$. The footrule and the overlap metrics are also indicated. As expected the third-order model ranks accurately the trails of length four and, we therefore use it as the reference ranking for the other models. The first-order model is able to identify three of the $top\_10$ trails although with a different ranking. The second-order model is able to identify eight of the $top\_10$ trails. For lower order models the trails in bold were identified from the reference ranking.

| ID | URL |
| --- | --- |
| 1 | / |
| 2 | /music/machines/analogue-heaven |
| 3 | /manufacturers/roland |
| 4 | /music/machines/samples.html |
| 5 | /music/machines |
| 6 | /samples.html |
| 7 | /machines |
| 8 | /manufacturers |
| 9 | /music/machines/categories/software/windows |
| 10 | /categories/software/windows |
| 11 | /music/machines/categories/drum-machines/samples |
| 12 | /analogue-heaven |
| 13 | /search.cgi |
| 14 | /guide |
| 15 | /analogue-heaven |
| 16 | /music/machines/images/farley3.jpg |
| 17 | /guide/finding.html |
| 18 | /categories/software/windows/readme |
| 19 | /categories/drum-machines/samples |
| 20 | /analogue-heaven/email.html |

Table 10: The URLs occurring in the rules given in Table 11

| rank | 1st Order | 2nd Order | 3rd Order |
|---|---|---|---|
| 1 | 2,15,1,F | **4,6,11,19** | 5,1,4,6 |
| 2 | 2,16,5,1 | **5,1,4,6** | 4,6,11,19 |
| 3 | **1,4,6,F** | **1,4,6,F** | 5,1,9,10 |
| 4 | 2,16,5,F | **5,1,9,10** | 5,1,12,3 |
| 5 | **5,1,4,6** | **7,1,9,10** | 1,4,6,F |
| 6 | 5,1,13,F | **4,6,4,6** | 4,6,4,6 |
| 7 | 2,15,1,8 | 1,9,10,F | 1,14,17,8 |
| 8 | 5,1,8,F | **5,1,8,3** | 5,1,8,3 |
| 9 | 2,15,20,F | **5,1,12,3** | 7,1,9,10 |
| 10 | **1,14,17,8** | 4,6,2,15 | 1,9,10,18 |
| footrule | 0.236 | 0.792 | 1.000 |
| overlap | 0.300 | 0.800 | 1.000 |

Table 11: The *top_10* rules for the MM data set obtained with models of first, second and third order for $\gamma = 0$

## 4 Concluding Remarks

In this work we have presented a model for implementing a variable length Markov chain (previosuly introduced in [BL05a, BL05b]) and a method for measuring the predictive power of such models (previously introduced in [BL06] using standard cross validation rather than the temporal-based version used herein). We then proposed a new method to evaluate the summarisation ability of the model, making use of the Spearman footrule metric to assess the accuracy with which a model represents the information content of a collection of user web navigation sessions.

A model that is able to accurately summarise the information content of a collection of user web navigation sessions can provide a platform for techniques focused on identifying user navigation patterns. Moreover, a model showing strong predictive power provides the means to predict the next link choice of unseen user navigation sessions and thus can be used for pre-fetching links or in adaptive web site applications.

We presented the results of an extensive experimental evaluation conducted on three real world data sets, which provide strong evidence that their is a linear relationship between predictive power and summarisation ability.

As future work, we are planning to incorporate our model within an adaptive web site platform in order to assess it as an online tool to provide link suggestions for users navigating within a site.